\documentclass{article}
\usepackage{spconf,amsmath,graphicx}

\usepackage{enumitem}
\setlist{nosep, leftmargin=14pt}

\usepackage[noadjust]{cite}
\usepackage{amssymb,amsfonts}
\usepackage{algorithmic}
\usepackage{textcomp}
\usepackage{xcolor}
\usepackage{booktabs} 
\usepackage{subcaption}
\usepackage{multirow}
\usepackage{pifont}
\usepackage[hidelinks]{hyperref}

\usepackage{algorithm}

\title{No Free Lunch in Annotation either: An objective evaluation of Foundation models for Streamlining Annotation in Animal Tracking}
\name{%
\begin{tabular}{cccc}
Emil Mededovic$^{\,1}$ & 
Valdy Laurentius$^{\,1}$ & 
Yuli Wu$^{\,1}$ & 
Marcin Kopaczka$^{\,1}$ \\
Zhu Chen$^{\,1}$ &
Mareike Schulz$^{\,2}$ &
René Tolba$^{\,2}$ &
Johannes Stegmaier$^{\,1}$\vspace{-0.3cm}
\end{tabular}}

\address{\normalsize $^{1}$ Institute of Imaging and Computer Vision, RWTH Aachen University, Aachen, Germany \\
\normalsize$^{2}$ Institute for Laboratory Animal Science, RWTH Aachen University, Germany\\
\normalsize E-mail: \texttt{\small\{emil.mededovic, johannes.stegmaier\}@lfb.rwth-aachen.de}}
\begin{document}

\maketitle

\begin{abstract}
We analyze the capabilities of foundation models addressing the tedious task of generating annotations for animal tracking. Annotating a large amount of data is vital and can be a make-or-break factor for the robustness of a tracking model. Robustness is particularly crucial in animal tracking, as accurate tracking over long time horizons is essential for capturing the behavior of animals. However, generating additional annotations using foundation models can be counterproductive, as the quality of the annotations is just as important. Poorly annotated data can introduce noise and inaccuracies, ultimately compromising the performance and accuracy of the trained model. Over-reliance on automated annotations without ensuring precision can lead to diminished results, making careful oversight and quality control essential in the annotation process. Ultimately, we demonstrate that a thoughtful combination of automated annotations and manually annotated data is a valuable strategy, yielding an IDF1 score of 80.8 against blind usage of SAM2 video with an IDF1 score of 65.6.
Our implementation for the annotation tool is available at: \url{https://github.com/medem23/SAM-QA}.
\end{abstract}

\section{Introduction}
\label{sec:intro}
Tracking systems that measure animal activity are crucial for assessing stress and severity indicators, providing invaluable insights into animal welfare \cite{Tolba, lauer2022multi}. To build these robust tracking models, a significant amount of annotated data is required. This demand underscores the need for an efficient, streamlined annotation process to support the development of reliable tracking systems capable of monitoring activity and welfare indicators over long periods.

In that regard, the introduction of foundation models has opened the door to streamlining annotation tasks, with the potential for increased speed and accuracy. The Segment Anything Model \cite{kirillov2023segment} is a promptable universal segmentation model designed for open-set segmentation. The prompts used for Segment Anything can include points, bounding boxes, or even masks, making it highly versatile for different types of segmentation tasks. In the field of generating annotations using foundation models, some preliminary work has already been published by \cite{archit2023segment}. In this study, the authors employed the Segment Anything Model (SAM) \cite{kirillov2023segment} to generate annotations for cells in microscopy images. They specifically followed the "annotate and fine-tune" approach, where SAM is used interactively to annotate the data, and then fine-tuned with the newly annotated data to improve its performance. The authors in \cite{sam-pt} combined the Segment Anything Model \cite{kirillov2023segment} with robust point trackers \cite{karaev2023cotracker, zheng2023point}, demonstrating that this integration results in a reliable video segmentation model \cite{sam-pt}. To accelerate the time-consuming annotation process, our SAM-QA (SAM Quality Annotation) approach leverages the concept of sequentially applying the Segment Anything Model (SAM) with automatically generated prompts. This technique forms the basis for streamlining high-quality annotations in video recordings of rodents. Our contribution introduces SAM-QA, an approach to streamline high-quality annotation production. We conducted a detailed analysis of SAM-QA on rodent datasets (rat and mouse), comparing it with classical segmentation techniques, watershed, open-set object detection, and recent Segment Anything Model extensions for video \cite{ravi2024sam}.





\begin{figure*}[t!]
\centering
\includegraphics[width=0.9\linewidth]{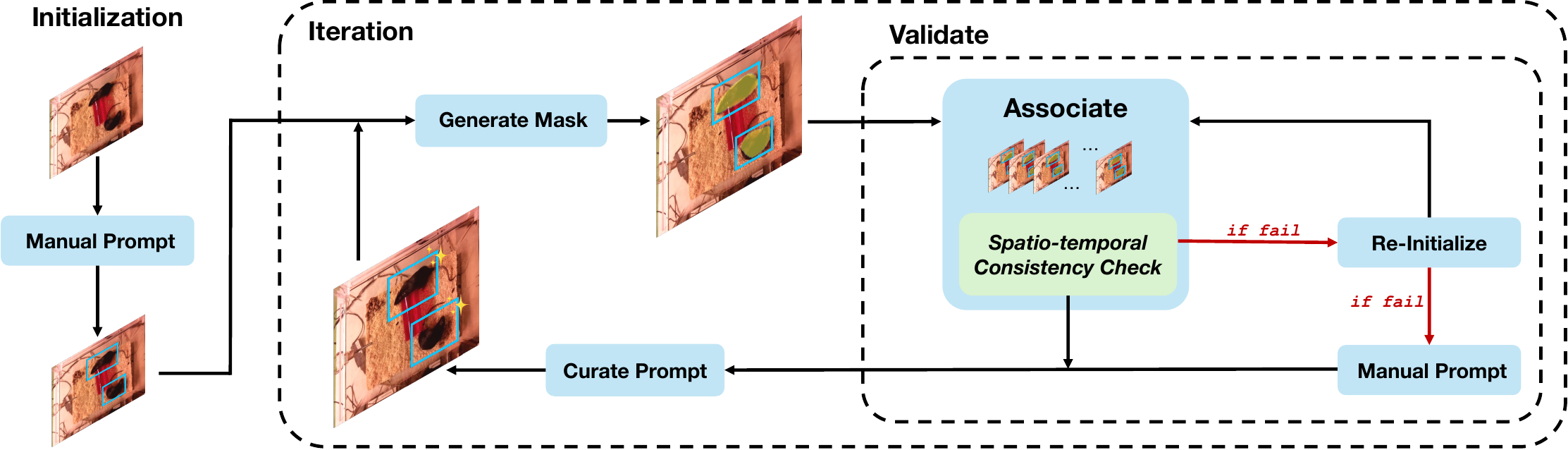}
\caption{The SAM-QA approach begins with the initialization step, where manual bounding box prompts are provided by the user. These prompts then enter an iterative loop where a fine-tuned and distilled SAM model generates segmentation masks. In the validation step, if the prompts are neither manual nor initial, an association check is performed to verify spatio-temporal consistency between the current and previous time steps. If these criteria are not met, a recovery attempt is made using SAM2 \cite{ravi2024sam}, which leverages equidistant grid-sampled point prompts. Should this recovery step also fail, the user is prompted to manually re-initialize. Finally, bounding boxes are generated from the validated masks, adjusted to account for rodent movement, and prepared for use in the subsequent time step.}
\label{fig:res}
\end{figure*}

\section{Methods}
\label{sec:Methodolgies}
In the following, we provide a detailed overview of the methods for semi-automatic data annotation applied and analyzed in our study, along with an in-depth description of the dataset and the tracker used.

\noindent\textbf{Dataset}. We have two datasets: a rat dataset with two rats consistently, and a mouse dataset featuring four mice. In Table \ref{tab:frame_counts},  the available training data and the lengths of the evaluation sequences are presented. The videos have a resolution of $1640\times 1232$ pixels and a frame rate of 30 frames per second.

\begin{table}[b]
\centering
\caption{Frame counts for training and evaluation sequences in rat and mouse datasets. Note that 9.0k frames correspond to a 5 minute sequence.}
\begin{tabular}{l c c c}
\toprule
\textbf{Dataset} & \textbf{Train Frames} & \textbf{Eval Seq 1} & \textbf{Eval Seq 2} \\ 
\midrule
Rats & 15.9k & 14.9k & 18.0k \\ 
Mice & 13.5k & 9.0k & 9.0k \\ 
\bottomrule
\end{tabular}
\label{tab:frame_counts}
\end{table}

\noindent \textbf{SAM-QA}. The basis of SAM-QA is a distilled and fine-tuned SAM \cite{kirillov2023segment}. Initially, we replace the base encoder with a smaller TinyViT model \cite{wu2022tinyvit} and perform knowledge distillation from the original SAM base encoder \cite{kirillov2023segment} using mean-squared error loss, as we do not require the general capability of SAM but instead focus on a narrow application with a static background. This modification allows us to employ a lightweight model, effectively speeding up inference. Subsequently, we fine-tune the lightweight SAM with TinyViT \cite{wu2022tinyvit} encoder on our animal datasets using cross-entropy and Dice loss \cite{sudre2017generalised}. The newly trained SAM is used at each time step to generate masks based on bounding box prompts. As illustrated in Fig. \ref{fig:res}, the SAM-QA approach consists of three stages: initialization, iteration, and a quality assessment validation step introduced at each iteration. In the initialization step, initial prompts are manually provided to begin the iteration process. These bounding box prompts are used to generate binary masks, which are then passed to the validation stage to assess mask quality. Before validation, the masks undergo a preprocessing step (Algorithm \ref{alg:remove_outliers}) to remove noise, as the rapid, non-linear movement of the rodents—sometimes even at 30 FPS—can introduce blur that leads to errors in the model's output. 
If a manual or initial mask is available, we assume it is of sufficient quality and proceed to the next frame. Otherwise, a spatio-temporal consistency check is conducted using masks from the previous time steps. 

Before assessing the spatio-temporal consistency, we begin by performing matching through an IoU-based association of masks between the current and previous frame. The matching criteria can now be formulated as follows:

\begin{enumerate}
    \item  \textbf{Overlap Condition}: For masks \( M_i \) and \( M_j \) in the same frame, the criterion is \textit{not met} if $\text{IoU}(M_i, M_j) > \beta$, where \( \beta \) is a user-defined threshold.
    \item \textbf{Size Condition}: Let \( A_Y \) be the area of the previous mask (manual prompt) and \( A_X \) the area of the current mask. The criterion is \textit{not met} if $A_X \notin [(1 - \alpha)A_Y, (1 + \alpha)A_Y]$.
\end{enumerate}




\noindent To pass the spatio-temporal consistency check, all conditions must be satisfied. If this check fails and re-initialization has not yet been performed, the method initiates automatic recovery using SAM2 \cite{ravi2024sam} with equidistant grid-sampled prompts, followed by a return to the association steps for re-evaluation. If this final attempt is unsuccessful, manual prompting becomes necessary; otherwise, the process advances to the next frame. We empirically select the following parameters: $\alpha=0.1$ and $\beta=0.9$.

\begin{algorithm}[t]
\caption{Remove Outliers in a Binary Mask by Isolating High-Density Regions}
\label{alg:remove_outliers}
\begin{algorithmic}[1]
    \REQUIRE Binary mask \texttt{mask} 
    \ENSURE Refined mask \texttt{mask\_wo\_outlier}
    \STATE \textbf{Extract $(x, y)$ coordinates:} Identify coordinates where \texttt{mask} $= 1$.
    \STATE \textbf{Density Estimation:} Stack coordinates as $[x, y]^T$ and apply Gaussian Kernel Density Estimation \cite{silverman2018density} to estimate density at each point:
    
$\qquad\displaystyle d(x, y) = \frac{1}{n h} \sum_{i=1}^{n} K \left( \frac{[x - x_i, y - y_i]^T}{h} \right), $

    where $K$ is the Gaussian kernel, $h$ (estimated using scott's rule \cite{scott2015multivariate}) is the bandwidth, and $n$ is the number of points.
    \STATE \textbf{Thresholding:} Set threshold at the 20th percentile, $\tau = \text{percentile}(d, 20)$. Retain points with $d(x, y) > \tau$.
    \STATE \textbf{Dilation:} Dilate with a fully-occupied $3 \times 3$ structuring element for 3 iterations to expand high-density regions.
\end{algorithmic}
\end{algorithm}

\begin{figure}[t!]
    \centering
    \includegraphics[width=0.97\linewidth]{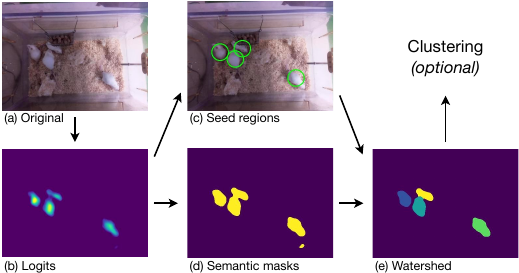}
    \caption{Illustration of the segmentation process using the watershed method: First, the image is passed through the segmentation model, and logits are used to identify peaks, which serve as seed points for watershed-based instance segmentation. Further refinement through clustering is optional.}
    \label{fig:2}
\end{figure}

\noindent\textbf{Segmentation \& Watershed}. To obtain bounding boxes using traditional methods, we train standard, widely-used segmentation models, such as U-Net \cite{ronneberger2015u} and DeepLabv3 \cite{chen2017rethinking}, all specifically for segmenting rodents. Additionally, we employ DINOv2, which remains frozen while an attached linear layer is trained \cite{oquab2024dinov}. All rat models are trained on 112 labeled images, while mouse models are trained on 113 labeled images. The training process utilizes a combination of cross-entropy and Dice loss functions \cite{sudre2017generalised}. The logit output from these models is then treated as a heatmap (Fig. \ref{fig:2}), from which we iteratively extract seed points by identifying peaks. Starting with the maximum intensity peak, additional peaks are sequentially added until the prior known number of animals is reached. For each selected peak, a circular exclusion zone proportional to the size of the animals is established to prevent detecting multiple peaks for the same rodent. If two exclusion zones overlap too much, the lower-intensity peak is removed. After peak identification, morphological closing and opening are performed to remove noise, and missed seeds are recovered based on prior knowledge of the expected number of elements in the image. 
Any outlier regions with areas disproportionately smaller than the largest current object are removed. We then apply the watershed algorithm \cite{beucher1979use} using inverted logits to generate instance segments. Finally, leveraging the prior knowledge of the number of elements, we employ clustering techniques such as $K$-means and Gaussian Mixture Models \cite{lloyd1982least, dempster1977maximum} to further refine the segmentation.

\noindent\textbf{Grounding DINO}. Grounding DINO \cite{liu2023grounding} is an open-set object detector. We prompt the model with either \texttt{Rat} or \texttt{Mice}, depending on the dataset, and then select the bounding boxes with the highest scores until reaching the total number expected (2 for rats, 4 for mice).

\noindent\textbf{Tracker}. For evaluation, we use the widely adopted ByteTrack tracker \cite{zhang2022bytetrack}, relying solely on IoU-based cost functions due to the visual similarity among rodents, which makes learning discriminative features challenging. We also fine-tune the tracker's association parameters to better suit our specific problem setting as follows:
\texttt{track\_high\_thresh = 0.5}, \texttt{track\_low\_thresh = 0.1}, \texttt{match\_thresh = 0.9}, \texttt{new\_track\_thresh = 0.9}, \texttt{track\_buffer = 120}. For the object detection model, we utilize the YOLOv8 model in its medium configuration \cite{yolov8_ultralytics}. Each model is trained for 10 epochs using SGD with a learning rate of 0.01, incorporating augmentations like hue adjustment, translation, scaling, flipping, and mosaic. All training sessions are conducted on a single NVIDIA RTX 3090.

\section{Results}
\label{sec:Results}
\begin{figure}[b!]
  \centering
  \begin{subfigure}{0.21\textwidth}
    \includegraphics[width=\textwidth]{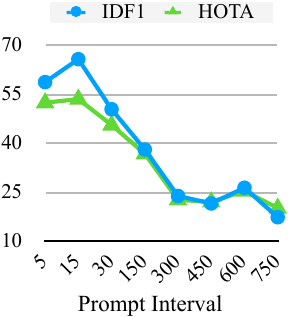}
    \caption{Rats}
    \label{fig:prompt_fre_rats}
  \end{subfigure}\hspace{1.5em}
  \begin{subfigure}{0.21\textwidth}
    \includegraphics[width=\textwidth]{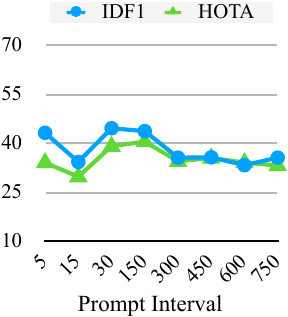}
    \caption{Mice}
    \label{fig:prompt_fre_mice}
  \end{subfigure}
  \caption{SAM2 video analysis for different prompt interval is presented. Prompt interval refers to the interval at which frames are manually annotated.}
  \label{fig:prompt_fre}
\end{figure}

\begin{table*}[t]
    \caption{This table presents the results for the downstream tracking task, evaluated using the well-established IDF1 \cite{ristani2016performance} and HOTA \cite{luiten2020IJCV} metrics (where higher scores indicate better performance). For post-processing, only the best-performing method is shown. Prompt Interval [Rats/Mice] refers to the interval at which frames are manually annotated. For SAM2 video (SAM-2V), we present the Prompt Interval for the top-performing results. Boldface results indicate the best performance, while underlined results represent the second-best performance, excluding manual annotation.}
    \centering
    \small
    \begin{tabular}{cccccccc}
    \toprule
      \multirow{2}{3em}[-0.2em]{\centering Model}   & \multirow{2}{6em}[-0.2em]{\centering Zero-Shot} & \multirow{2}{8em}[-0.2em]{\centering Post-processing} & \multirow{2}{6em}[-0.2em]{\centering Prompt Interval}  & \multicolumn{2}{c}{Rats} &  \multicolumn{2}{c}{Mice} \\ \cmidrule(lr{0.4em}){5-6} \cmidrule(l{0.4em}r){7-8}
      & &  & & {\centering IDF1 $\uparrow$}  & {\centering HOTA $\uparrow$} & {\centering IDF1 $\uparrow$}  & {\centering HOTA $\uparrow$} \\\midrule
        U-Net \cite{ronneberger2015u} & \ding{55}  & GMM \cite{dempster1977maximum} & \ding{55} & 57.7 & 44.9 & 28.1 & 24.1 \\
        DeepLabv3 \cite{chen2017rethinking} & \ding{55}  & GMM \cite{dempster1977maximum} & \ding{55} & 60.2 & 44.7 & 20.5 & 20.1 \\
        DINOv2 \cite{oquab2024dinov}  & \ding{55} & K-Means \cite{lloyd1982least} & \ding{55} & 52.6 & 40.9 & 14.7 & 12.8 \\
        U-Net \cite{ronneberger2015u}  & \ding{55}  & GMM \cite{dempster1977maximum}& 15/10 & 45.7 & 40.9 & 21.6 & 21.2 \\
        \midrule
          Grounding DINO \cite{liu2023grounding}   & \ding{51} & \ding{55} &  \ding{55} & 37.0 & 31.0 & 29.8 & 26.8 \\
          Grounding DINO \cite{liu2023grounding} & \ding{51} & \ding{55} &  15/10 & 23.9 & 24.9 & \underline{45.1} & 37.4 \\
        \midrule
        SAM-2V \cite{ravi2024sam}  & \ding{51} & \ding{55} & 15/15 & \underline{65.6} & \underline{53.4} & 34.2 & 29.7 \\
        SAM-2V \cite{ravi2024sam}  & \ding{51} & \ding{55} & 30/30 & 50.3 & 45.5 & 44.5 & \underline{39.0} \\\midrule
         SAM-QA (ours)  & Distilled \& Finetuned  & \ding{55} & 21/9 & \textbf{80.8} & \textbf{59.1} & \textbf{61.1} & \textbf{46.0} \\
        \midrule
       \multicolumn{4}{c}{\textbf{Manual Annotation}} & 85.0 & 80.9 & 76.6 & 59.0 \\\bottomrule
    \end{tabular}
    \label{tab:result}
\end{table*}
In Table \ref{tab:result}, we present the downstream tracking results using newly generated training labels across various methods. Performance is evaluated using IDF1 \cite{ristani2016performance} and HOTA \cite{luiten2020IJCV} metrics. As our objective involves assessing animal activity to infer behavioral and severity patterns, we focus primarily on IDF1 score to test long-term association accuracy. 

Our first observation reveals a notable performance discrepancy between rats and mice. For manually annotated labels, the difference is less pronounced but still evident, suggesting that tracking smaller, more numerous objects with low contrast against the background (as in the mice dataset) is inherently challenging (Fig. \ref{fig:2}). This performance gap is especially prominent in segmentation followed by watershed approaches, highlighting these approaches are not suited for too complex configurations.

When comparing methods, we find that Grounding DINO \cite{liu2023grounding}, as a zero-shot object detector, is suboptimal for generating training bounding boxes, as expected. For a fair comparison, we conducted an analysis where we combined manual and generated annotations across traditional methods and Grounding DINO \cite{liu2023grounding}. Interestingly, this approach yielded divergent results: for rats, the model performed worse, while for mice, performance improved. This discrepancy can be attributed to the visibility and ambiguity of tails in rat data, leading to inconsistent bounding box creation (e.g., varying tail visibility). For rats, this mix yielded only 23.9 IDF1—a drop of 13.1—though HOTA scores decreased less, suggesting that while detections were more frequent, they introduced considerable tracking ambiguities. For mice, however, combining manual annotations improved IDF1 by 15.3, slightly outperforming SAM-2V \cite{ravi2024sam}, likely due to reduced tail visibility and therefore more consistent bounding box generation.

Traditional methods based on semantic segmentation models followed by watershed segmentation perform well as long as the dataset complexity is moderate and overlapping instances are limited. For example, on the rats dataset, the best-performing model achieves an IDF1 score of 60.2, only 5.4 shy of SAM-2V \cite{ravi2024sam} results. However, SAM-2V \cite{ravi2024sam} demonstrates strong and promising performance in zero-shot settings, only underperforming compared to SAM-QA, which uses a fine-tuned SAM and robust quality assessment. In SAM-QA, prompts are not set at fixed intervals but are triggered by conflict occurrences, enhancing accuracy. Since SAM-2V \cite{ravi2024sam} is not fine-tuned, ambiguities arise, potentially impacting results as illustrated in Fig. \ref{fig:prompt_fre}. For the Mice dataset, overall performance is suboptimal, likely due to the challenging nature of the task, where even a small amount of noisy labels can break the entire tracking model (Fig. \ref{fig:prompt_fre_mice}). The model consistently maintains its score regardless of prompt frequency, indicating promising potential for performance improvements through fine-tuning. Nevertheless, SAM-2V’s \cite{ravi2024sam} lack of fine-tuning leads to challenges in handling occlusions, especially when rats (Fig. \ref{sec:Methodolgies}) enter or exit the transparent red tunnel (toy for enrichment). As we increase the interval between prompts (Fig. \ref{fig:prompt_fre_rats}), performance decreases, with IDF1 scores dropping from 65.6 to 17.3 at a prompting interval of 750 (25 seconds).

Our proposed annotation tool, SAM-QA, outperforms other methods, improving IDF1 by 15.2 for rats and 16 for mice, with a comparable number of interventions to the second-best performing approach based on SAM-2V \cite{ravi2024sam}. It is important to note that this performance boost benefits significantly from fine-tuning and targeted intervention in specific conflict cases via our consistency check. Nonetheless, while our method yielded the best results among all tested approaches, it still lags behind fully manual annotations, trailing by around 4.2 IDF1 for rats and 15.5 for mice. These findings underscore the importance of caution when using foundation models for label generation, as they may produce weak labels that lack the precision of manual annotations. 

\section{Conclusion}
We demonstrate the need for caution when applying foundation models for label generation in tracking tasks, given the precision required to maintain track consistency. Inconsistent bounding boxes can lead to suboptimal model performance due to noise. We propose an iterative approach that integrates a lightweight, fine-tuned Segment Anything Model (SAM) with a quality assessment process to ensure label consistency, benefiting the tracking task. While our approach currently falls short of manual annotation in accuracy, SAM-2V shows promising results and will be a focus of future research, aiming for seamless integration with our quality assessment process and potential further fine-tuning on our dataset.  


\section{COMPLIANCE WITH ETHICAL STANDARDS}
Ethical approval was not required as the videos were obtained without direct interaction or handling of the animals. The recordings were sourced from previous studies, ensuring no additional impact on the animals during this work.

\section{Acknowledgments}
\label{sec:acknowledgments}
This work was funded by the German Research Foundation DFG with the grants STE2802/4-1 (EM) and STE2802/5-1 (ZC).

\bibliographystyle{IEEEbib}
\bibliography{refs}

\end{document}